\documentclass[letterpaper, 10 pt, conference]{ieeeconf} 
\IEEEoverridecommandlockouts                           

\usepackage{graphics} 
\usepackage{epsfig} 
\usepackage{mathptmx} 
\usepackage[T1]{fontenc}
\usepackage{amsmath} 

\usepackage{amssymb}  
\usepackage{siunitx}
\usepackage{cite}
\makeatletter
\let\NAT@parse\undefined
\makeatother

\usepackage{hyperref}

\title{\LARGE \bf
Control-Oriented Learning for Dynamic Tracking and Stability Analysis of Soft Pneumatic Actuators}
\begin{document}

\author{Nithin S. Kumar$^{1}$ and Eric J. Barth$^{1}$
\thanks{$^{1}$Department of Mechanical Engineering, Vanderbilt University, Nashville TN 37215}
\thanks{\tt\small Email: nithin.s.kumar@vanderbilt.edu}
\thanks{This work has been submitted to the IEEE for possible publication. Copyright may be transferred without notice, after which this version may no longer be accessible.}}
\maketitle
\thispagestyle{empty}
\pagestyle{empty}

\begin{abstract}

Soft pneumatic actuators offer inherent compliance and safe interaction but remain difficult to model and control because of their highly nonlinear, distributed dynamics. We present a control-oriented data-driven modeling and control framework that decomposes actuator behavior into a nonlinear static equilibrium model and a linear residual dynamics model identified using Extended Dynamic Mode Decomposition with control (EDMDc). This representation enables feedforward compensation, task-space feedback control, and local closed-loop stability analysis through an augmented linear model. Experiments achieve $\approx$1 mm root mean square error (RMSE) during low-speed ($\approx$10 mm/s) trajectory tracking and below 10 mm RMSE at higher speeds ($\approx$100 mm/s). The framework further achieves stable tracking of highly dynamic user-generated references with peak accelerations exceeding 25 m/s$^2$ while simultaneously performing real-time obstacle avoidance. Finally, the proposed stability analysis is experimentally validated by accurately predicting stable, marginal, and unstable operating regimes. These results demonstrate that structured, control-oriented learning provides an accurate and practical framework for soft actuator control.

\end{abstract}

\section{INTRODUCTION}

Soft continuum robots have attracted significant attention because their intrinsic compliance enables safe interaction with humans, delicate objects, and uncertain environments. Unlike conventional rigid-link manipulators, soft robots undergo large continuous deformations, making them well suited for applications such as minimally invasive surgery, rehabilitation, wearable robotics, industrial manipulation, and exploration of confined or unstructured environments \cite{Rodrigue2024,Shintake2018,Calisti2017}.

Despite these advantages, the same compliance that enables these capabilities also makes soft robots difficult to model and control. Soft actuators exhibit effectively infinite-dimensional kinematics, strong geometric nonlinearities, hysteresis, and parameter variations arising from fabrication and material aging \cite{Xavier2022,Tawk2021}. Consequently, achieving accurate dynamic task-space control remains a fundamental challenge, particularly beyond quasi-static operation \cite{DellaSantina2023}.

To address this challenge, considerable effort has been devoted to developing dynamic models for soft robot control. Reduced-order continuum formulations, including piecewise constant curvature (PCC) \cite{Katzschmann2019}, finite element \cite{Duriez2013}, Cosserat rod \cite{Till2019}, and discrete elastic rod models \cite{Rucker2022} provide physically interpretable descriptions that form the basis of model-based control strategies. However, these approaches require substantial analytical effort, actuator-specific parameter identification, and simplifying assumptions. As actuator complexity increases, maintaining both model fidelity and real-time computational efficiency becomes increasingly difficult.

Alternatively, data-driven methods, including Gaussian processes \cite{Fang2019}, Koopman-based methods \cite{Haggerty2023}, sparse identification \cite{Papageorgiou2024}, and reinforcement learning \cite{Thuruthel2019} learn system behavior directly from experimental data without explicit first-principles modeling. However, many learning-based approaches behave as black-box input-output models, require large training datasets, or provide limited physical interpretability and theoretical guarantees for feedback control. 

These observations suggest that neither purely physics-based nor purely data-driven approaches fully satisfy the requirements of practical soft actuator control. Physics-based methods provide structure and interpretability but often rely on restrictive assumptions, whereas purely learned methods reduce modeling effort at the expense of analytical structure. A promising alternative is therefore a control-oriented learning framework that combines the strengths of both approaches.

\begin{figure}[t]
  \centering
  \includegraphics[width=\columnwidth]{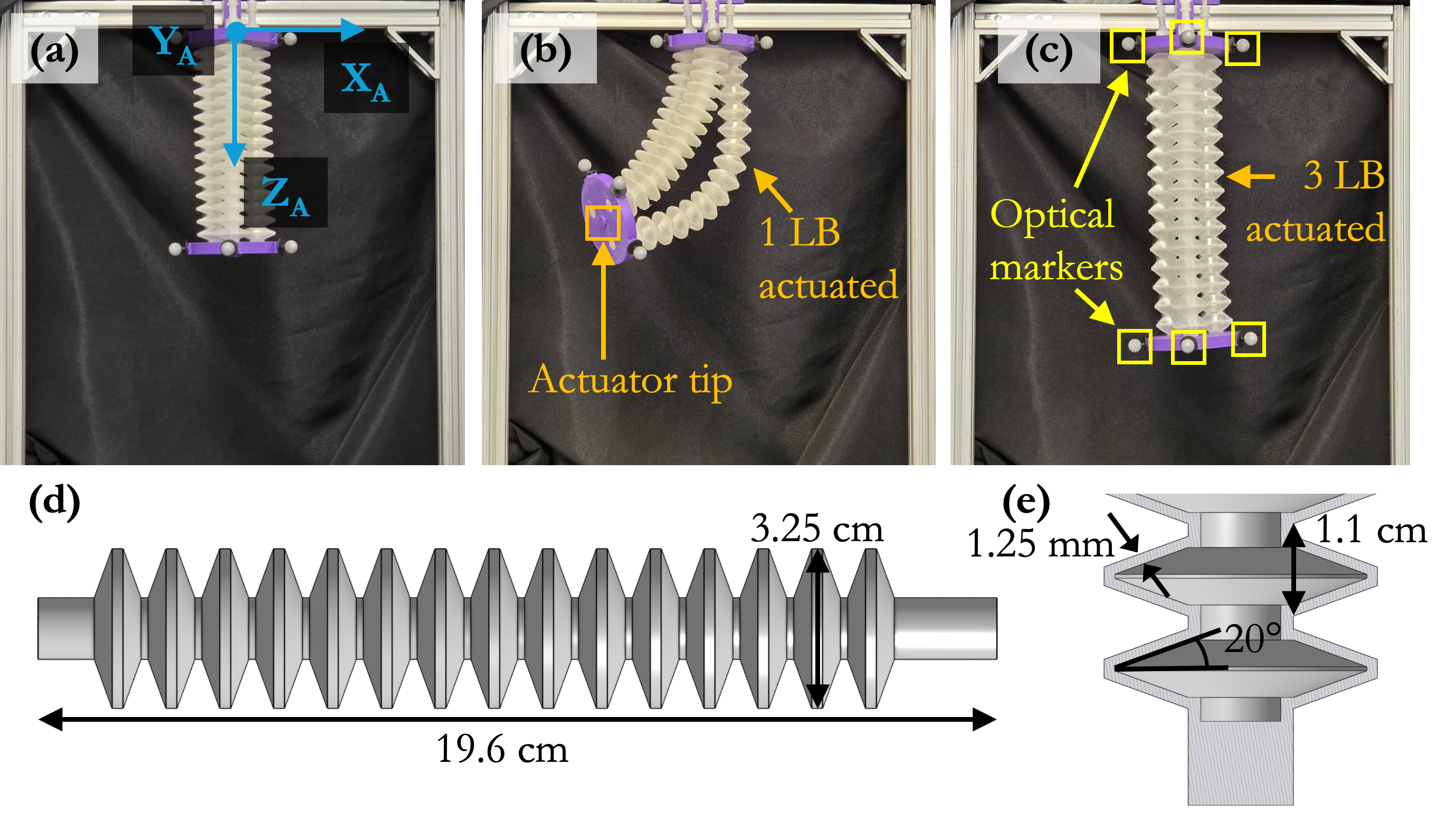}
  \caption{Spatial soft pneumatic actuator comprised of three linear bellows (LB) and experimental platform. (a) Actuator-fixed coordinate frame $\{X_A,Y_A,Z_A\}$ defined by the three base optical markers. (b) Example deformation produced by independent actuation of a single bellow. (c) Linear extension produced by simultaneous actuation of all three bellows together with the six optical tracking markers used for tip position estimation. (d) Bellows geometry and overall actuator dimensions. (e) Cross-sectional view of a single compliant bellows segment.}
  \label{fig:actuator}
\end{figure}

In this paper, we present a control-oriented data-driven modeling and control framework for soft pneumatic actuators. Rather than learning the complete nonlinear dynamics, the proposed approach combines a learned nonlinear static equilibrium model with a linear residual dynamics model identified using Extended Dynamic Mode Decomposition with control (EDMDc) \cite{Proctor2016,Williams2015}. The resulting representation enables task-space feedback control and local stability analysis without requiring explicit first-principles modeling. The proposed framework is experimentally validated on a soft pneumatic actuator (Fig. \ref{fig:actuator}) through a series of trajectory-tracking, interactive-motion, and stability experiments.

The primary contributions of this work are:
\begin{itemize}
    \item A control-oriented learned representation of a continuum actuator that preserves the simplicity of linear feedback design while capturing dominant nonlinear behavior.
    \item Experimental validation of accurate trajectory tracking, including stable tracking of highly dynamic (>25~m/s$^2$) user-generated references and real-time obstacle avoidance.
    \item A model-based local stability analysis that predicts stable, marginal, and unstable closed-loop operating regimes prior to hardware deployment.
\end{itemize}

\section{HARDWARE} \label{sec:hardware}

\subsection{Pneumatic soft actuator}

The experimental platform is a spatial soft pneumatic actuator (Fig.~\ref{fig:actuator}) consisting of three parallel linear bellows connecting a rigid base plate and tip plate. Independent pressurization of each bellow produces axial elongation (Fig.~\ref{fig:actuator}c), while differential pressurization generates three-dimensional bending of the actuator (Fig.~\ref{fig:actuator}a).

Each bellow was fabricated using a Bambu Lab X1C 3D printer with Polymaker PolyFlex TPU 90A and a uniform wall thickness of 1.25~mm. The bellows comprise 15 compliant triangular segments and have an overall actuator length of 196~mm with a segment diameter of 32.5~mm. The base and tip plates were 3D printed from PLA (Bambu Lab PLA Basic) and serve both as structural interfaces for the bellows and as mounting fixtures for three optical tracking markers each. The base plate is rigidly attached to an external frame.

The actuator is used as a representative soft robotic platform for evaluating the proposed modeling and control framework rather than as an application-specific device. Its combination of large deformation, strongly nonlinear pressure-to-position behavior, and coupled pneumatic dynamics provides a challenging yet experimentally tractable benchmark for data-driven modeling and control.

\subsection{Experimental setup}

The actuator is driven by a custom pneumatic control box that independently regulates the pressure supplied to each bellow through 4~mm ID / 6~mm OD pneumatic tubing. The actuator is mounted to an external frame constructed from T-slotted aluminum rails.

The control box houses three 5-port/3-way proportional spool valves (Festo MPYE-5-M5-010-B) together with in-line pressure sensors (Festo SDE5-D10-NF-T14-V-M8). A mixed I/O data acquisition board (Humusoft MF634) interfaces the pneumatic hardware to a desktop PC (AMD Ryzen 7 5700G, 3.80~GHz) running MATLAB R2024b Simulink Desktop Real-Time, enabling independent closed-loop pressure regulation of each bellow.

\begin{figure}[t]
  \centering
  \includegraphics[width=\columnwidth]{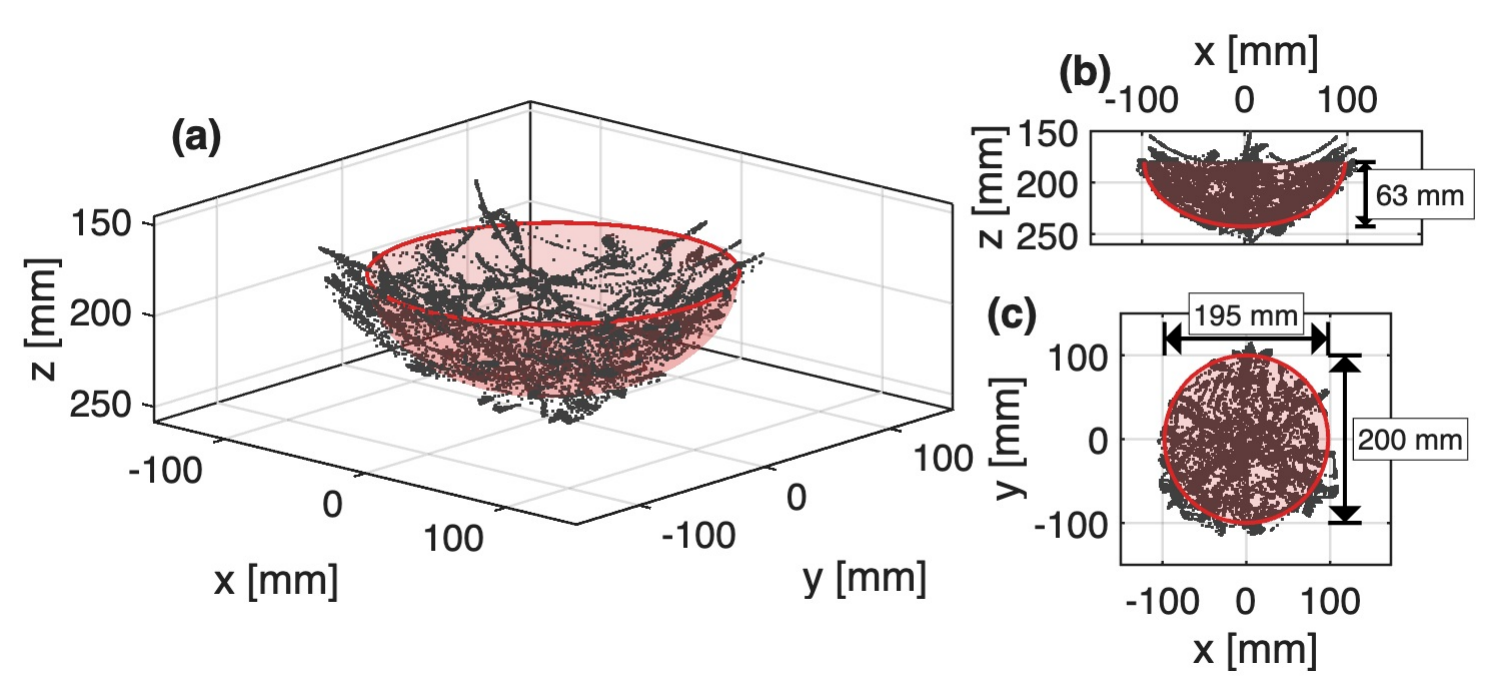}
  \caption{Reachable workspace identified from the training dataset together with the best-fit half-ellipsoid used for real-time workspace projection. (a) Three-dimensional workspace. (b) x-z projection showing the workspace height. (c) x-y projection showing the workspace footprint. The fitted ellipsoid is used to project user references outside the reachable workspace during interactive motion generation.}
  \label{fig:wkspace}
\end{figure}

The actuator is instrumented with six optical markers. Three markers mounted on the base plate define the actuator-fixed coordinate frame $\{X_A,Y_A,Z_A\}$ (Fig.~\ref{fig:actuator}a), while three markers mounted on the tip plate are used to compute the actuator tip position in this local frame. Marker positions are measured using an OptiTrack Prime~13 motion capture system providing Cartesian position measurements at 50~Hz. Fig.~\ref{fig:wkspace} shows the experimentally identified reachable workspace together with its least-squares best-fit half-ellipsoidal approximation (of dimensions 195 mm in $x$, 200 mm in $y$, and 63 mm in $z$). This workspace approximation is later used to project user-generated references onto the reachable workspace.

Motion capture, feedforward computation, and feedback control are executed at 50~Hz, while pressure sensing and low-level pressure control operate at 1~kHz. 

\section{MODEL AND CONTROLLER}
\label{sec:model_control}

The proposed framework decomposes the actuator behavior into a nonlinear static component that captures the equilibrium input-output relationship and a linear residual dynamic component that models transient deviations from equilibrium. This decomposition simplifies controller design by confining the learned dynamics to a linear residual model while allowing the nonlinear static behavior to be compensated through feedforward actuation.

\begin{figure}[t]
  \centering
  \includegraphics[width=\columnwidth]{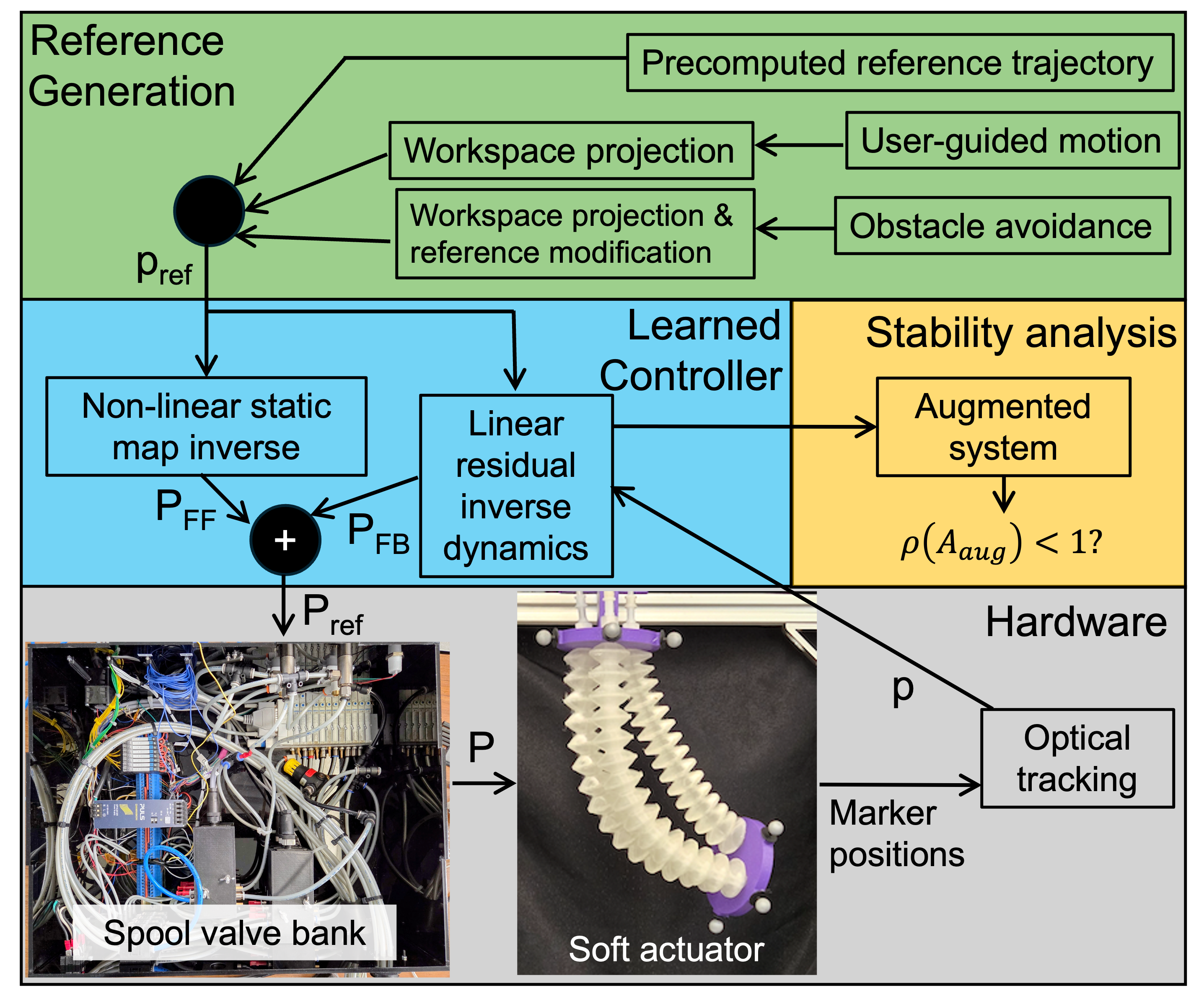}
  \caption{Overview of the proposed control-oriented learning framework. Different reference-generation modules (reference trajectories, user-guided motion, and obstacle avoidance) provide the desired tip position, $p_{\mathrm{ref}}$. The learned control core generates the equilibrium pressure $P_{\mathrm{FF}}$ from the static model and the feedback pressure $P_{\mathrm{FB}}$ from the residual dynamic model, yielding the desired pressure command $P_{\mathrm{ref}} = P_{\mathrm{FF}} + P_{\mathrm{FB}}$. A low-level pressure controller regulates the chamber pressures $P$, producing the actuator tip position $p$, which is fed back to the controller. The learned residual model also enables local closed-loop stability analysis.}
  \label{fig:system}
\end{figure}

The control input is the chamber pressure vector:
\[
P=[P_1,P_2,P_3]^T\in\mathbb{R}^{3 \times 1},
\]
where \(P_i\) denotes the regulated pressure supplied to the \(i^{\text{th}}\) bellow. The measured output is the actuator tip position expressed in the actuator-fixed frame:
\[
p=[x_{\mathrm{tip}},\,y_{\mathrm{tip}},\,z_{\mathrm{tip}}]^T\in\mathbb{R}^{3 \times 1},
\]
obtained from the optical motion capture system. 
Figure~\ref{fig:system} provides an overview of the proposed control-oriented learning framework. The learned static mapping, residual dynamic model, and task-space feedback controller together generate the actuator pressure commands, while the same residual model also supports local stability analysis. The following subsections describe each component in detail.

\subsection{Static Input-Output Mapping}

The equilibrium behavior of the actuator is represented by a nonlinear static mapping from chamber pressure to actuator tip position,
\begin{equation}
    p \approx W^T\Theta(P),
\end{equation}
where $W\in\mathbb{R}^{q\times3}$ is the learned coefficient matrix and $\Theta(P)\in\mathbb{R}^{q \times 1}$ is a polynomial basis evaluated at the current pressure input. A second-order polynomial basis was selected to capture the dominant nonlinear pressure-to-position relationship while maintaining computational efficiency leading to:
\begin{align}
\Theta(P)= [&1,\;
P_1,\;
P_2,\;
P_3,\;
P_1^2,\;
P_2^2,\;
P_3^2,\;
P_1P_2,\;
P_1P_3,\;
P_2P_3]^T,
\end{align}
resulting in $q=10$ basis functions. Second-order polynomial bases have also been identified as the dominant terms using sparse regression for soft pneumatic actuators in \cite{Papageorgiou2024}. 

\begin{figure}[t]
  \centering
  \includegraphics[width=\columnwidth]{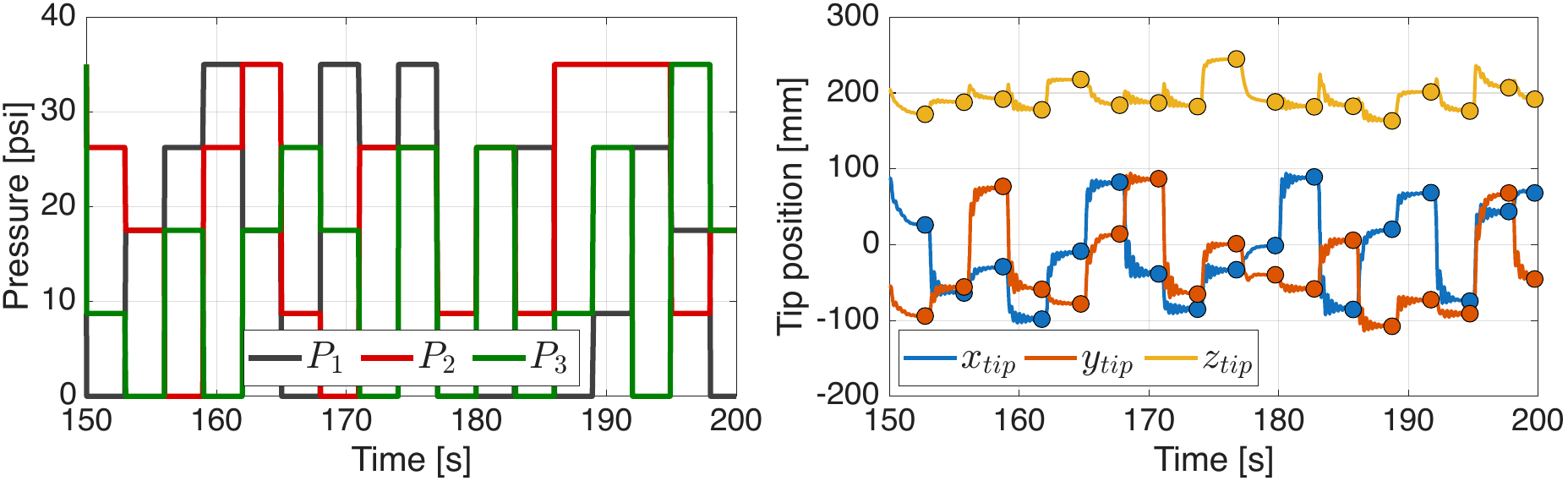}
  \caption{Representative 50 s segment of the staircase training dataset used to identify the static input-output mapping. (Left) Independent chamber pressure excitations. (Right) Corresponding actuator tip position. Equilibrium samples (circles) are obtained by averaging the final 0.5 s of each 3 s hold period before fitting the static polynomial model.}
  \label{fig:training}
\end{figure}

The model is identified from steady-state experimental data collected using staircase pressure inputs (between 0-35 psi/0-241 kPa) that uniformly excite the actuator workspace. The static model is identified from 500~s of staircase pressure excitations, yielding 161 equilibrium pressure-position pairs spanning the actuator workspace shown in Fig.~\ref{fig:wkspace}. A 50 s segment of the training data is shown in Fig.~\ref{fig:training}. Each pressure command ($P_{t_i}$) is held for 3 s, allowing the actuator to converge to equilibrium, and the corresponding tip position ($p_{t_i}$) is computed as the mean of the final 0.5 s of each hold period. We stack the basis evaluations and measured tip positions to generate the matrices:
\begin{align}
    \Phi&=[\Theta(P_{t_1})^T;\cdots;\Theta(P_{t_m})^T] \in \mathbb{R}^{m\times q}, \\
    Y&=[p_{t_1}^T;\cdots;p_{t_m}^T] \in \mathbb{R}^{m\times3},
\end{align}
to find $W$ such that $Y\approx \Phi W$. This is obtained by regularized least squares:
\begin{equation}
W=(\Phi^T\Phi+\lambda I)^{-1}\Phi^TY,
\end{equation}
where $\lambda=\num{1e-4}$ is the Tikhonov regularization parameter.

Table~\ref{table_static_map} summarizes the static-map fitting error for polynomial degrees $d=1$--$4$. Although higher-order models further reduce the fitting error, they substantially increase the number of basis terms ($q=20,35$ for $d=3,4$, respectively). Since the remaining modeling error is compensated by the residual feedback controller, the $d=2$ model provides an effective balance between model fidelity, interpretability, and real-time implementation.

For a desired reference position $p_{\mathrm{ref},k}$ at time $t_k$, the corresponding feedforward pressure command $P_{\mathrm{FF},k}$ is obtained by numerically inverting the learned static map. At each update, the bounded optimization problem:
\begin{equation}
\min_{P}\;
\|W^T\Theta(P)-p_{\mathrm{ref},k}\|_2^2
+\lambda_P\|P\|_2^2
+\lambda_{\Delta P}\|P-P_{k-1}\|_2^2
\end{equation}
is solved subject to:
\[
P_{\min}\le P\le P_{\max}.
\]
The parameters $\lambda_P=10,\text{ }\lambda_{\Delta P}=10$ were selected empirically to balance tracking accuracy, control effort, and input smoothness. 

\begin{table}[b]
\caption{Static Map RMSE for Polynomial Fit Degrees $d$}
\label{table_static_map}
\begin{center}
\begin{tabular}{|c||c|c|c|c|}
\hline
\textbf{Degree} $d$ & \multicolumn{4}{c|}{\textbf{RMSE [mm]}}\\
\cline{2-5}
& $x_{tip}$ & $y_{tip}$ & $z_{tip}$ & Total\\
\hline\hline
1 & $9.12$ & $9.30$ & $17.37$ & $21.71$ \\
\hline
2 & $7.94$ & $8.34$ & $5.79$ & $12.89$ \\
\hline
3 & $6.16$ & $6.60$ & $5.61$ & $10.63$ \\
\hline
4 & $5.00$ & $5.72$ & $5.62$ & $9.45$ \\
\hline
\end{tabular}
\end{center}
\end{table}

The resulting feedforward command $P_{\mathrm{FF}}(t)$ approximates the equilibrium pressure required to achieve the desired tip position, while the residual feedback controller described below compensates the remaining transient and modeling errors. For the precomputed reference trajectory tracking experiments, the static map is inverted pointwise using MATLAB's \textit{fmincon}. For the user-guided experiments, real-time operation is achieved using a warm-started Jacobian-based Gauss--Newton solver with the analytic Jacobian of the polynomial model and backtracking line search.

\subsection{Residual Dynamic Model}

Using the same 500 s staircase dataset, the feedforward tip position $p_{\mathrm{FF}}$ is first computed from the learned static model. Unlike the static mapping, which uses only the 161 equilibrium pressure-position pairs, the residual dynamics are identified from the complete 500 s dataset. The residual at time $t_k$ is defined as:
\begin{equation}
r_k=p_k-p_{\mathrm{FF},k}\in\mathbb{R}^{3\times 1}.
\end{equation}

The residual therefore represents the transient deviation from the static equilibrium prediction. The residual state is augmented using time-delay coordinates:
\begin{equation}
z_k=
[r_k,\;
r_{k-1},\;
\cdots,\;
r_{k-t_r}]^T
\in\mathbb{R}^{3(t_r+1)\times 1},
\end{equation}
where $t_r$ denotes the residual history length. Throughout this work, $t_r=2$ was found to capture the dominant actuator and pneumatic dynamics while maintaining a compact lifted-state representation. The lifted-state snapshots are assembled as (let $n_z=3(t_r+1)$ and $n_u=3$):
\begin{align}
Z_1 &= [z_1,\ldots,z_K]\in\mathbb{R}^{n_z \times K},\\
Z_2 &= [z_2,\ldots,z_{K+1}]\in\mathbb{R}^{n_z \times K},\\
U &= [P_{t_1},\ldots,P_{t_K}]\in\mathbb{R}^{n_u \times K},
\end{align}
where the inputs and lifted states are normalized using their training-set mean ($\mu_Z \in \mathbb{R}^{n_z\times 1},\mu_U \in \mathbb{R}^{n_u\times 1}$) and standard deviation ($S_z \in \mathbb{R}^{n_z\times n_z},S_u \in \mathbb{R}^{n_u\times n_u}$) yielding $\tilde{Z}_1,\tilde{Z}_2$, and $\tilde{U}$.

The residual dynamics are modeled as the discrete-time linear system:
\begin{equation}
\tilde{z}_{k+1}
=
A\tilde{z}_k
+
B\tilde{U}_k,
\label{eq:edmdc}
\end{equation}
where $A$ and $B$ are identified using EDMDc. Defining:
\begin{align}
\Omega=
\begin{bmatrix}
\tilde{Z}_1\\
\tilde{U}
\end{bmatrix},
\qquad
G=
\begin{bmatrix}
A&B
\end{bmatrix},
\end{align}
we solve $\tilde{Z}_2=G\Omega$ using regularized least squares:
\begin{equation}
G
=
\tilde{Z}_2
\Omega^T
(\Omega\Omega^T+\lambda I)^{-1},
\end{equation}
where $\lambda=\num{1e-4}$. The output matrix $C = [I_3 \quad0]$ extracts the current residual from the lifted state. The predicted tip position is then reconstructed as:
\begin{equation}
\hat{p}_{k+1}
=
p_{\mathrm{FF},k+1}
+
\hat{r}_{k+1},
\end{equation}
where $\hat{r}_{k+1}=C\hat{z}_{k+1}$ after reversing the normalization. Although this prediction can be used for one-step-ahead estimation, feedback is computed directly from the measured tip position. The residual model is used instead for state propagation and local stability analysis in Section \ref{sec:results_stab}.

\begin{figure}[t]
  \centering
  \includegraphics[width=\columnwidth]{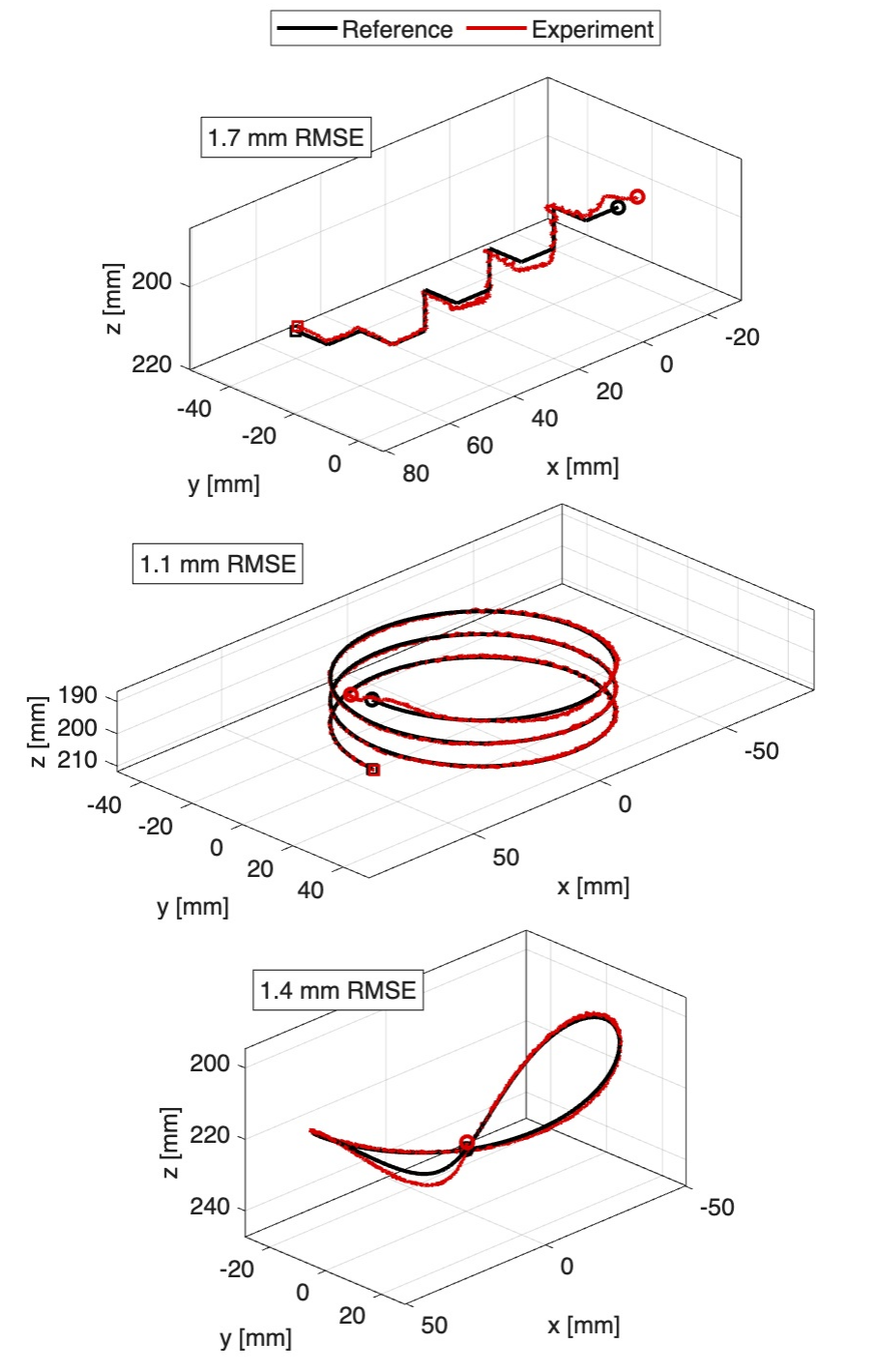}
  \caption{Reference trajectory tracking task. Desired (black) and measured (red) tip trajectories for the staircase (6.5 mm/s), helix (10 mm/s), and lemniscate (10 mm/s) references. Trajectories begin at $\circ$ and end at $\square$.}
  \label{fig:reftrack}
\end{figure}

\subsection{Controller}

To determine the feedback pressure $P_{\mathrm{FB}}$, we consider the mapping from pressure to residual space (i.e., $\Delta r = M\Delta P$) and compute the corresponding pressure correction using its regularized pseudo-inverse, $\Delta P_{\mathrm{FB}}=M^T(MM^T + \alpha I)^{-1}\Delta r_{\mathrm{des}}$.

The mapping $M\in \mathbb{R}^{3 \times 3}$ is obtained directly from the learned residual model. Since the normalized lifted dynamics satisfy $\Delta\tilde{z}=B\Delta\tilde{P}$, denormalization yields $\Delta z=S_zBS_u^{-1}\Delta P$. Using $\Delta r=C\Delta z$ gives:
\begin{equation}
M=CS_zBS_u^{-1}.
\end{equation}

Let the tracking error be $e=p_{\mathrm{ref}}-p$ and the integral error be $E=\int e\,dt$. The desired residual correction is generated by the task-space PI controller:
\begin{equation}
\Delta r_{\mathrm{des}}
=
K_Pe
+
K_IE,
\end{equation}
where $K_P,K_I\in\mathbb{R}^{3\times3}$ are diagonal gain matrices. The resulting feedback pressure is therefore:
\begin{equation} \label{eq:fb}
P_{\mathrm{FB},k}
=
M^T(MM^T+\alpha I)^{-1}
\left(K_Pe_k+K_IE_k\right),
\end{equation}
where $\alpha=\num{1e-4}$ is the Tikhonov regularization parameter. The final pressure command is:
\begin{equation}
P_k=P_{\mathrm{FF},k}+P_{\mathrm{FB},k},
\end{equation}
which is saturated to the allowable pressure limits (0-35 psi/ 241 kPa) before being sent to the pressure valve command. 

\section{RESULTS} \label{sec:results}

This section experimentally validates the proposed framework on the spatial soft pneumatic actuator (Fig. \ref{fig:actuator}). We first establish the controller's baseline closed-loop tracking accuracy using representative reference trajectories, then demonstrate real-time, highly dynamic interactive behaviors, and finally compare the experimentally observed closed-loop behavior with the proposed stability analysis. The same controller gains, $K_P=\mathrm{diag(2,2,2)}$ and $K_I=\mathrm{diag(5,5,5)}$, were used for all precomputed reference and user-guided tracking experiments.

\begin{figure}[t]
  \centering
  \includegraphics[width=\columnwidth]{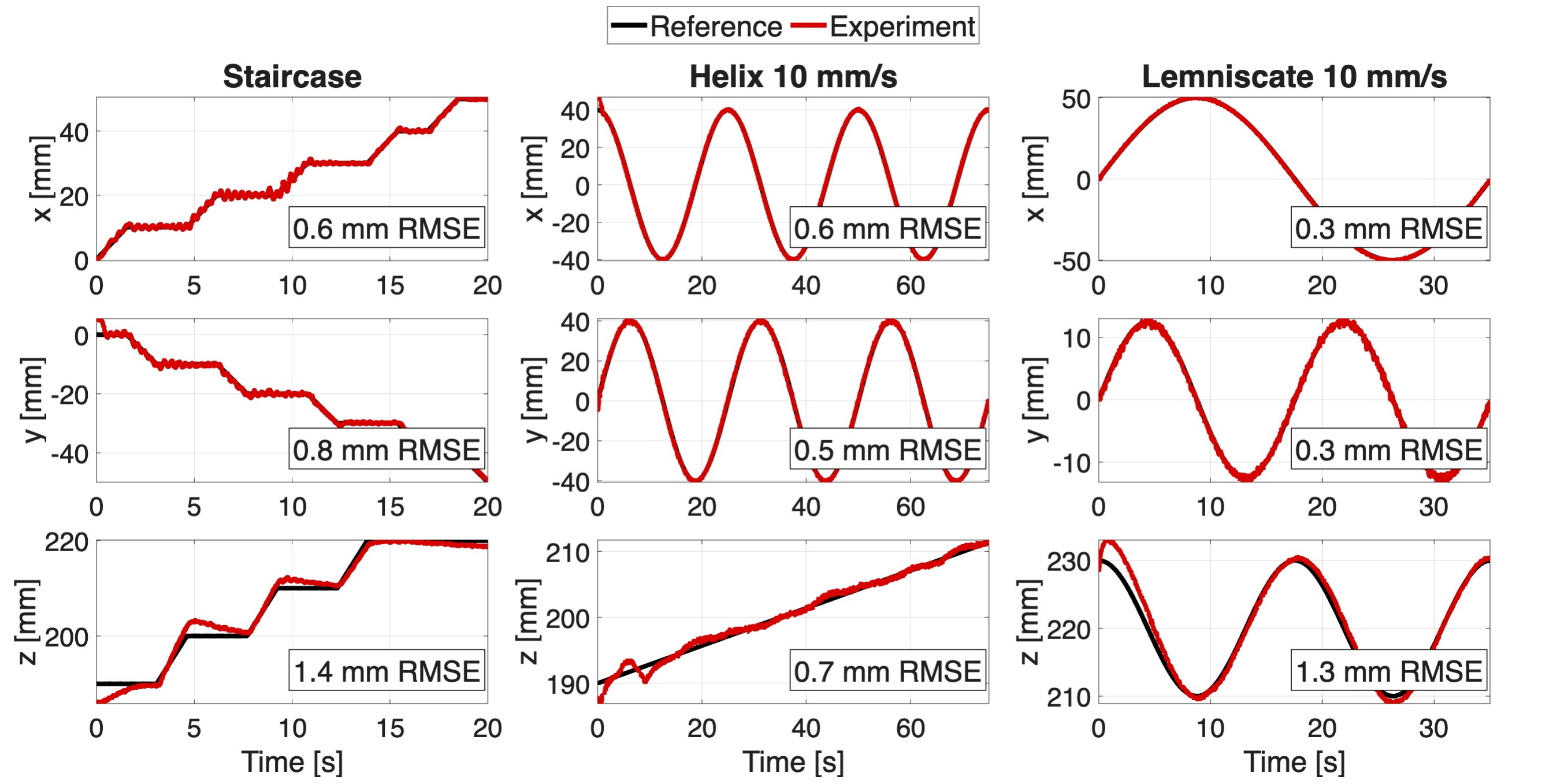}
  \caption{Cartesian position histories corresponding to the trajectories in Fig. \ref{fig:reftrack}. The proposed controller accurately tracks the desired motion in all three coordinates while maintaining smooth transient behavior. Reported RMSE values correspond to each Cartesian component.}
  \label{fig:reftrack_time}
\end{figure}

\subsection{Reference trajectory tracking}

The controller was first evaluated on three reference trajectories expressed in task-space. The staircase trajectory evaluates independent linear motion along the $x$, $y$, and $z$ directions, the helix assesses coupled circular motion with simultaneous axial translation, and the lemniscate continuously excites all three Cartesian coordinates simultaneously. Collectively, these trajectories evaluate independent axis and coupled spatial motion with continuously varying curvature, providing a broad assessment of controller performance.

Figure~\ref{fig:reftrack} compares the desired and measured tip trajectories in task space, while the corresponding Cartesian position histories are shown in Fig.~\ref{fig:reftrack_time}. In all cases, the actuator accurately reproduces the prescribed three-dimensional motion with little steady-state error and no observable oscillatory behavior.

\begin{table}[h]
\caption{Reference Tracking Performance}
\label{table_rmse}
\begin{center}
\begin{tabular}{|l||c|c|}
\hline
\textbf{Trajectory} & \textbf{Ref. Speed [mm/s]} & \textbf{Avg. RMSE [mm]}\\
\hline\hline
Helix       & 10           & $1.09$ \\
\hline
Helix       & 100           & $9.47$ \\
\hline
Lemniscate  & 10           & $1.36$ \\
\hline
Lemniscate  & 100           & $6.60$ \\
\hline
Staircase   & 6.5           & $1.79$ \\
\hline
\end{tabular}
\end{center}
\end{table}

Quantitative tracking performance is summarized in Table~\ref{table_rmse}, where the reported RMSE values are averaged over three trials. 

\subsection{Dynamic User-Generated Trajectory Tracking}

Next, we evaluate the framework in a real-time interactive setting. A handheld optical marker is used to generate reference positions in real-time, providing a practical benchmark for highly dynamic, user-guided motion. At each control update, the user marker is expressed in the actuator-fixed frame and projected onto the fitted workspace ellipsoid (Fig.~\ref{fig:wkspace}) whenever it falls outside the actuator's reachable workspace. The projected position is then supplied directly to the controller. 

\begin{figure}[t]
  \centering
  \includegraphics[width=\columnwidth]{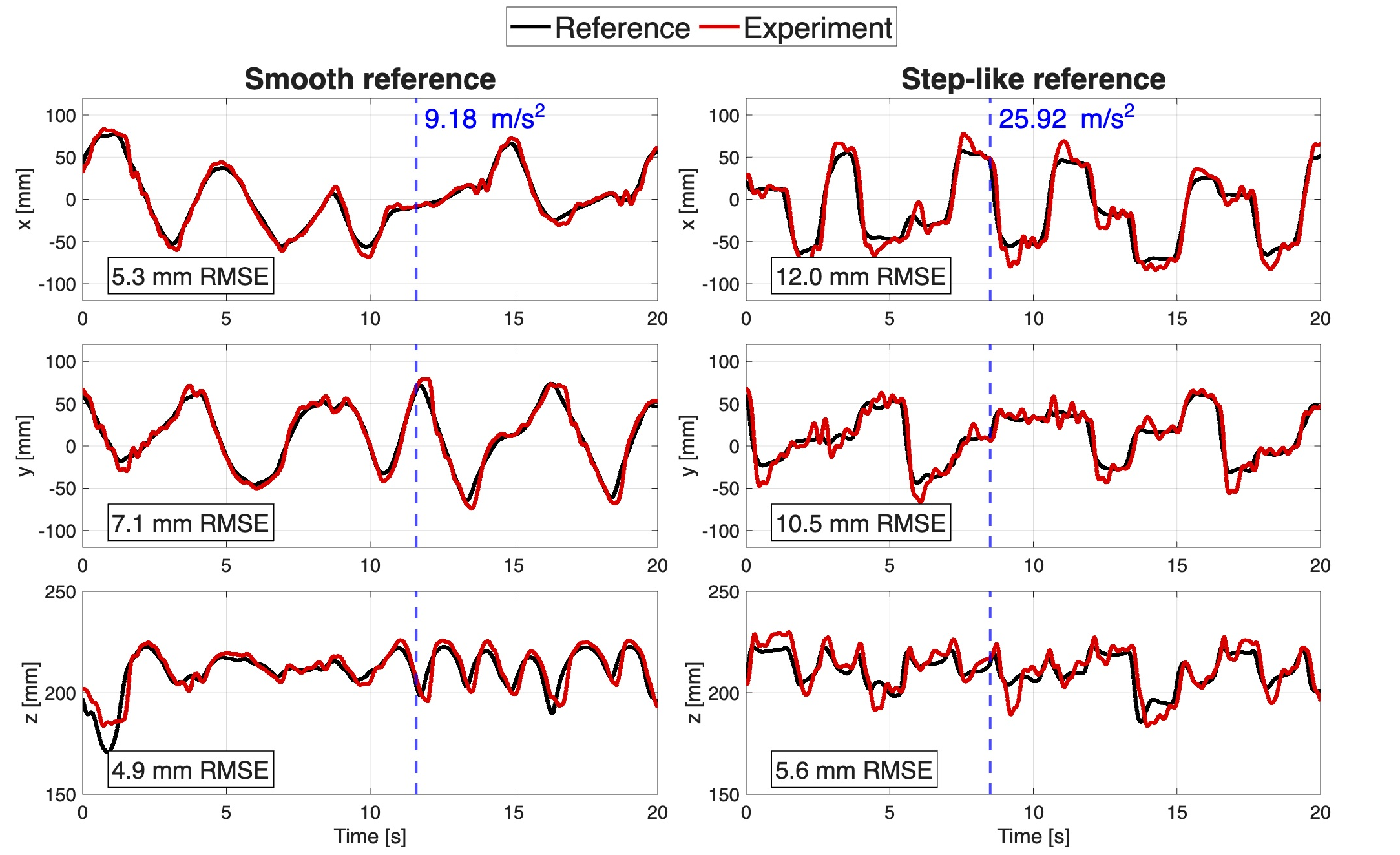}
  \caption{Real-time user-guided trajectory tracking. Smooth (left) and step-like (right) references generated from a handheld optical marker together with the measured actuator response. The smooth reference resulted in a peak reference acceleration and overall RMSE of 9.18 m/s$^2$ and 10.1 mm, respectively. The corresponding values for the step reference were 25.92 m/s$^2$ and 16.8 mm.}
  \label{fig:usertrack_time}
\end{figure}

\begin{figure}[b]
  \centering
  \includegraphics[width=\columnwidth]{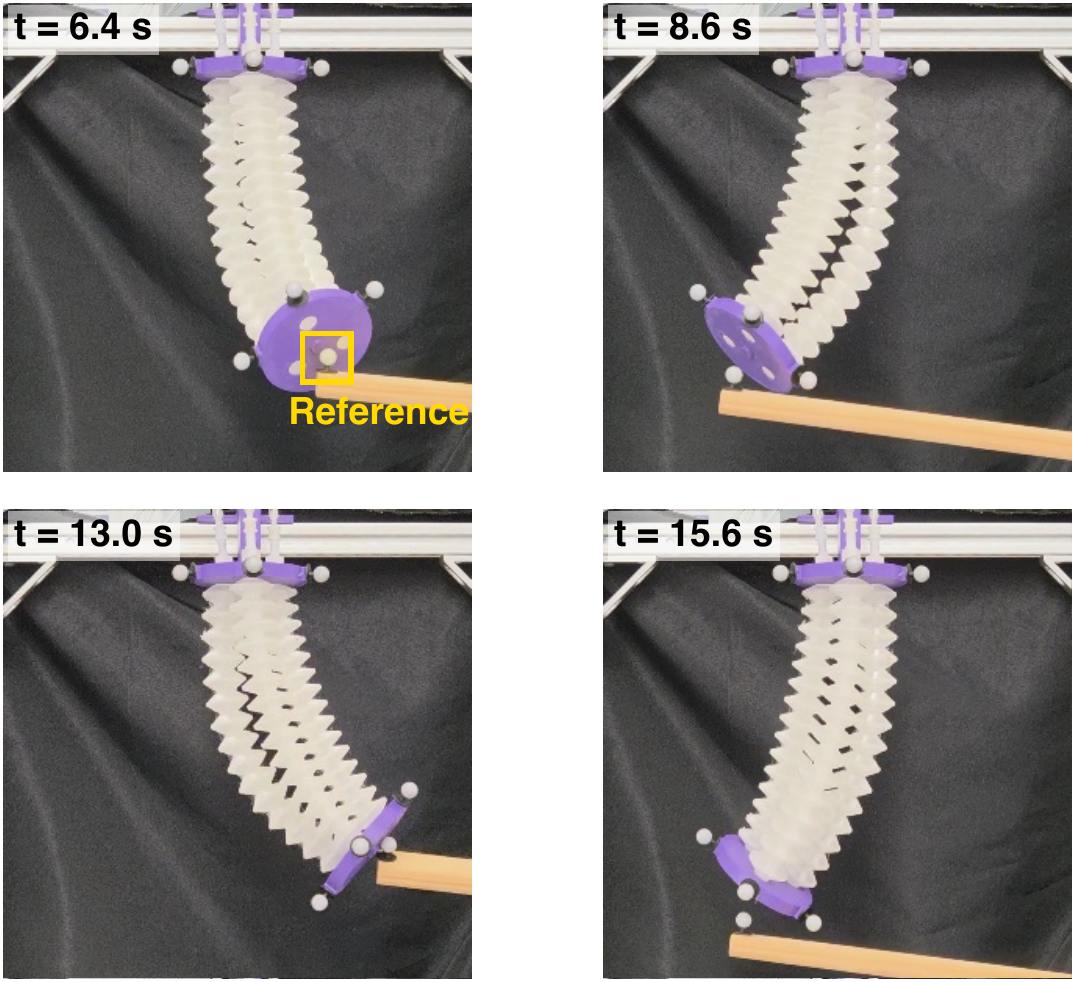}
  \caption{Representative snapshots of the interactive user-tracking experiment. The desired reference is generated in real-time from the handheld optical marker while the actuator tracks the projected reference. Please refer to the supplemental materials for a video.}
  \label{fig:usertrack_pics}
\end{figure}

To evaluate the controller over a range of user inputs, two representative motions were considered: smooth continuous motion and deliberately aggressive, step-like motion. Figure~\ref{fig:usertrack_time} presents representative smooth (left) and step-like (right) user-guided motions. The corresponding peak reference accelerations are 9.18~m/s$^2$ and 25.92~m/s$^2$ with $\approx$1 cm and <2 cm RMSE, respectively, demonstrating that the proposed framework is capable of tracking highly dynamic user commands while maintaining stable closed-loop behavior. Representative snapshots of this experiment are shown in Fig.~\ref{fig:usertrack_pics}. 

To demonstrate that the controller is independent of the reference-generation strategy, obstacle avoidance is implemented as a real-time reference-modification layer without altering the underlying controller.

Let $p_o(t)\in\mathbb{R}^{3\times1}$ denote the time-varying obstacle position and $p(t)\in\mathbb{R}^{3\times1}$ the measured tip position. Obstacle avoidance is activated whenever the obstacle-tip Euclidean distance is within a prescribed outer safety distance, $r_{\mathrm{out}}=100$~mm. Define $L(t)=\|p(t)-p_o(t)\|$ and  the unit vector $\hat{n}(t)=\frac{p-p_o}{L}$, which points from the obstacle toward the tip. The corresponding safety reference is $p_{\mathrm{safe}}(t)=p_o+r_{\mathrm{in}}\hat{n}$, where $r_{\mathrm{in}}=25$~mm is the prescribed minimum safety distance.

For $r_{\mathrm{in}}<L<r_{\mathrm{out}}$, the modified reference is obtained by linearly blending the nominal and safety references:
\begin{equation}
p_{\mathrm{mod}}(t)
=
(1-w)p_{\mathrm{ref}}
+
wp_{\mathrm{safe}},
\end{equation}
where $w=\frac{r_{\mathrm{out}}-L}{r_{\mathrm{out}}-r_{\mathrm{in}}}$. When $L\le r_{\mathrm{in}}$, the reference is set to $p_{\mathrm{safe}}$, whereas for $L\ge r_{\mathrm{out}}$ the nominal reference is recovered. The modified reference is checked if it is beyond the actuator's workspace (and if so, projected onto the reachable ellipsoidal workspace) before being supplied to the controller.

\begin{figure}[t]
  \centering
  \includegraphics[width=\columnwidth]{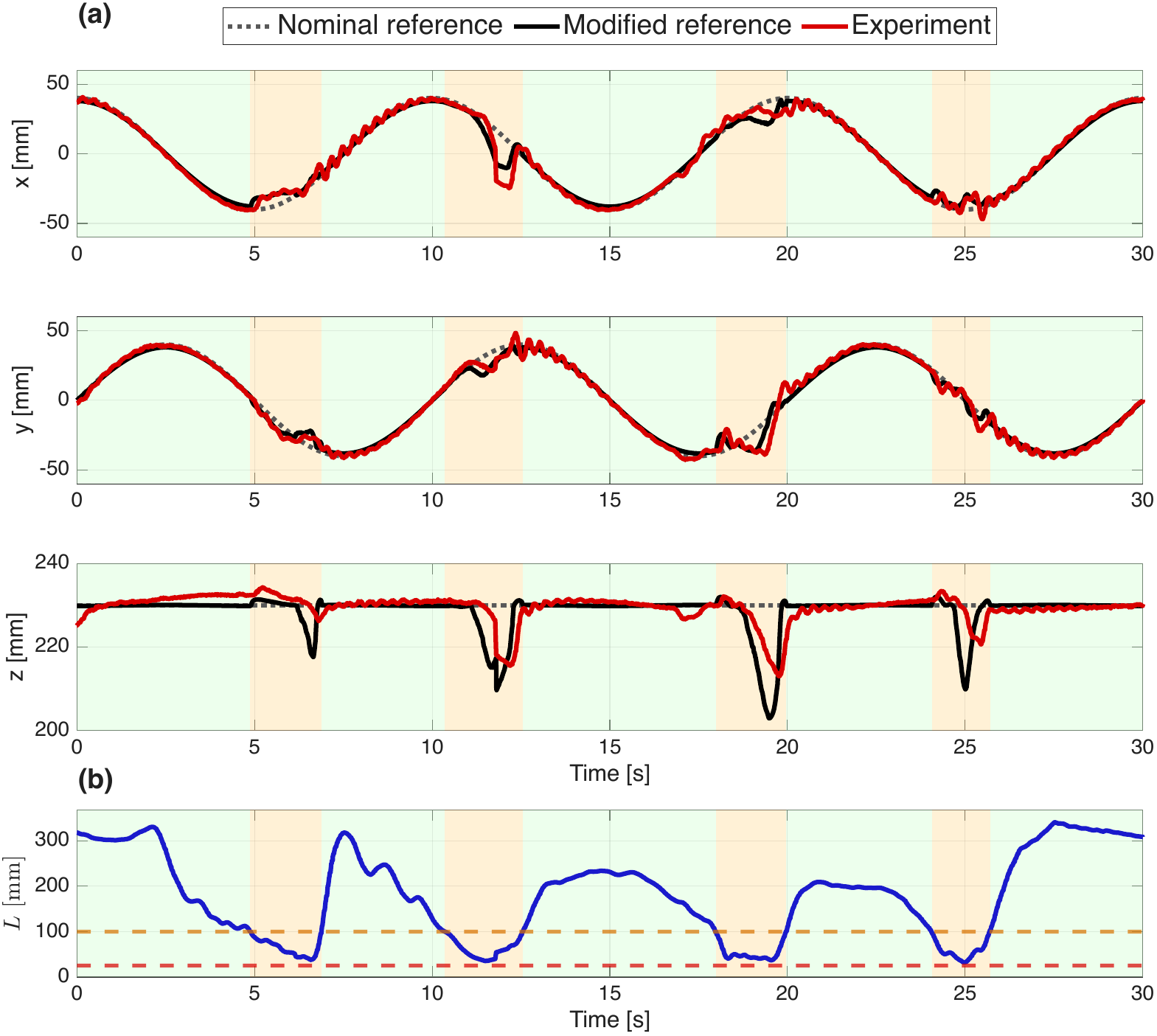}
  \caption{Dynamic obstacle avoidance during tracking of a 0.1 Hz circular reference. (a) Cartesian position histories. Orange shaded regions indicate intervals when the obstacle is within 100 mm of the actuator tip (reference modification is active), whereas green regions correspond to nominal tracking. (b) Obstacle-tip Euclidean distance ($L$). The dashed orange and red lines denote the outer (100 mm) and inner safety distances (25 mm), respectively. The proposed reference modification maintains the prescribed minimum separation while returning smoothly to the nominal reference once the obstacle-tip distance exceeds 100 mm.}
  \label{fig:obs_time}
\end{figure}

Figure~\ref{fig:obs_time}a shows the resulting reference modification and actuator response. The user-held marker now acts as a dynamic obstacle (i.e., $p_o(t)$) to avoid while the actuator tip simultaneously tracks a 0.1~Hz circular reference in the $X_A-Y_A$ plane. The nominal trajectory is temporarily displaced (four times in Fig. \ref{fig:obs_time}) to maintain a safe distance from the moving obstacle before smoothly returning to the original reference once the obstacle is sufficiently far away (i.e., $L>r_{\mathrm{out}}$). Figure~\ref{fig:obs_time}b plots $L(t)$ during this experiment. The horizontal lines indicate $r_{\textrm{in}}$ and $r_{\textrm{out}}$ while the orange shaded regions denote intervals during which the reference modification is active (i.e., $L<r_{\textrm{out}}$), while the green regions indicate nominal tracking. Throughout the experiment, the controller maintained the prescribed minimum separation ($r_{in}$) while smoothly transitioning between the nominal and modified references without instability or discontinuities.

\begin{figure}[t]
  \centering
  \includegraphics[width=\columnwidth]{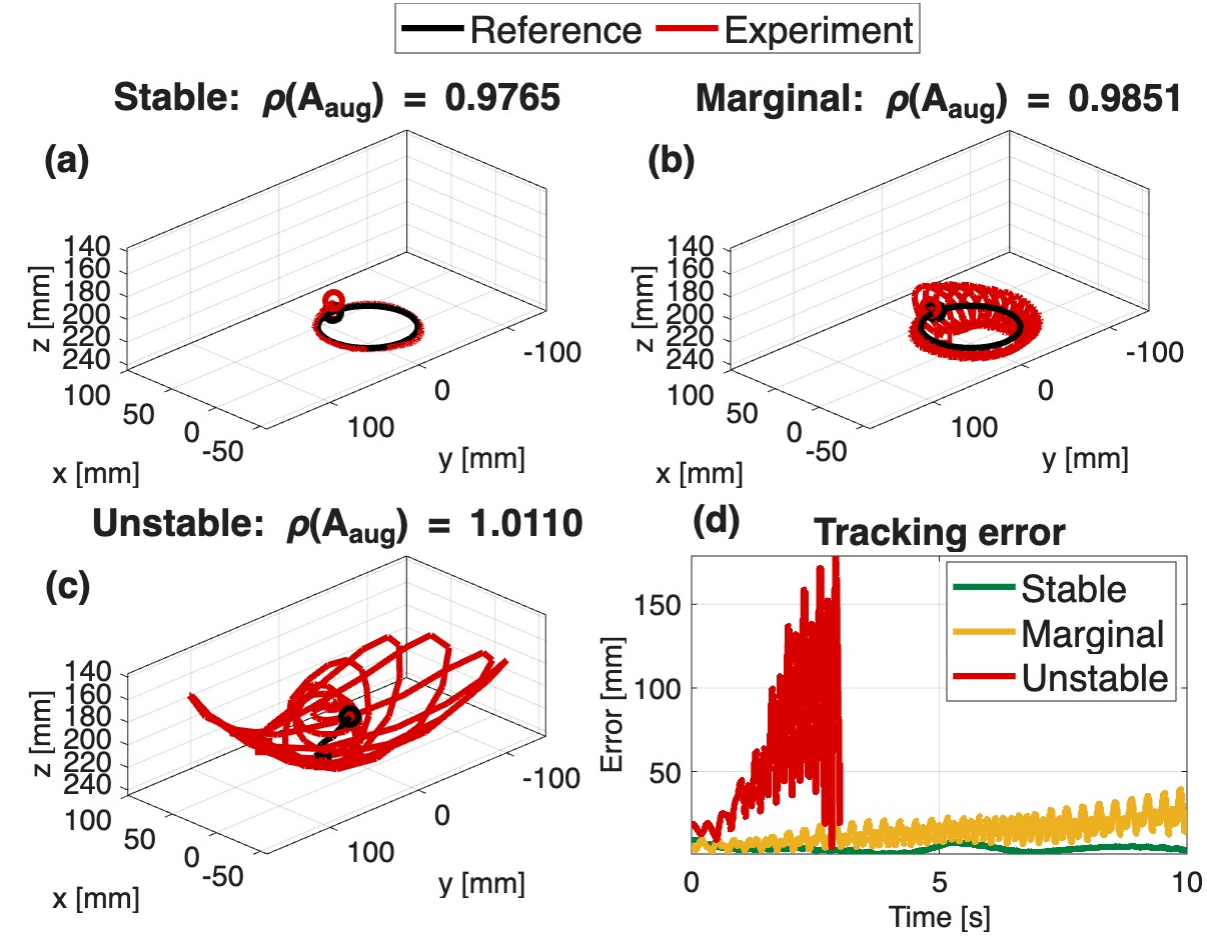}
  \caption{Experimental validation of the proposed stability analysis. Representative closed-loop responses for (a) stable ($\rho(A_{aug})<1$), (b) marginal ($\rho(A_{aug})\approx1$), and (c) unstable ($\rho(A_{aug})>1$) operating regimes during 0.1 Hz circular tracking with randomly sampled feedback gains. (d) The corresponding tracking-error (Euclidean distance) histories illustrate the transition from stable convergence to divergent behavior.}
  \label{fig:stability}
\end{figure}

\subsection{Stability Analysis} \label{sec:results_stab}

The proposed framework also enables local closed-loop stability analysis, allowing candidate feedback gains to be screened prior to hardware experiments. By combining the learned residual dynamics with the feedback controller, we derive the augmented closed-loop system dynamics and use the resulting augmented state transition matrix $A_{\mathrm{aug}}$ to predict local closed-loop stability. Stability is predicted whenever the spectral radius of $A_{\mathrm{aug}}$ satisfies $\rho(A_{\mathrm{aug}})<1$. For reference, the identified nominal residual model has $\rho(A)=0.9540$, indicating stable open-loop residual dynamics. A complete derivation is provided in the Appendix~\ref{sec:appendix}. This criterion characterizes the local stability of the learned linearized closed-loop model about the operating point and should not be interpreted as a guarantee of global nonlinear stability.

The proposed stability criterion was evaluated by randomly perturbing all entries of the feedback gain matrices $K_P, K_I$ (using MATLAB's \textit{randn}), constructing the corresponding augmented system, computing $\rho(A_{aug})$, and executing the resulting controller on hardware. Figure~\ref{fig:stability} summarizes representative responses during a 0.1~Hz circular tracking task. Gain matrices with dominant magnitudes of $K_P \sim 2, K_I \sim 10$ produced $\rho(A_{aug})<1$, resulting in stable tracking (Fig. \ref{fig:stability}a). Gains $K_P \sim 5, K_I \sim 5$ resulted in $\rho(A_{aug})$ approaching 1 leading to oscillatory behavior (Fig. \ref{fig:stability}b). Finally, gain matrices with dominant magnitudes of $K_P \sim 10, K_I \sim 1$ yielded $\rho(A_{aug})>1$ leading to the closed-loop response diverging rapidly and requiring early termination to prevent actuator damage (Fig. \ref{fig:stability}c). These results demonstrate that the learned augmented model reliably predicts local closed-loop behavior across stable, marginal, and unstable operating regimes. 

\section{DISCUSSION} \label{sec:discussion}

The proposed control-oriented modeling framework demonstrates that accurate dynamic control of soft pneumatic actuators can be achieved without sacrificing analytical tractability. We discuss its implications for stability analysis, closed-loop performance, and practical limitations.

\subsection{Stability analysis as a design tool}

A key advantage of the proposed control-oriented modeling framework is that it enables local closed-loop stability analysis directly from the learned dynamics. Unlike many highly expressive data-driven models, the residual representation admits an augmented linear closed-loop system whose spectral radius can be evaluated prior to hardware deployment, reducing reliance on empirical trial-and-error tuning.

Although the proposed criterion does not constitute a global nonlinear stability proof, it provides a practical engineering tool for distinguishing stability regimes. This is particularly valuable for pneumatic soft actuators, where delays, valve nonlinearities, and viscoelastic effects can rapidly destabilize aggressive feedback controllers.

\subsection{Reference generation and tracking performance}

A practical advantage of the proposed framework is the explicit separation between reference generation and low-level control. The same learned static map, residual dynamics model, and PI feedback controller are used for reference trajectory tracking, interactive user following, and obstacle avoidance. Different behaviors are realized solely by modifying the reference supplied to the controller. In particular, obstacle avoidance is achieved by geometrically modifying the nominal reference in real-time rather than altering the underlying control law. This structure naturally accommodates additional reference-generation modules, such as perception- or planning-based motion generation, without requiring controller redesign.

The ablation study in Table~\ref{table_ablation} highlights the complementary roles of the static and dynamic components. Feedback alone outperforms feedforward alone, indicating that accurate closed-loop regulation is more important than precise equilibrium compensation. Combining both, however, consistently yields the best performance.

\begin{table}[h]
\caption{Ablation study comparing static feedforward (FF), residual feedback (FB), and the complete controller (FF+FB)}
\label{table_ablation}
\begin{center}
\begin{tabular}{|l|c||c|c|c|}
\hline
\textbf{Trajectory} & \textbf{Ref. Speed} & \multicolumn{3}{c|}{\textbf{RMSE [mm]}}\\
\cline{3-5}
                     & [mm/s]         & FF only & FB only & FF+FB\\
\hline\hline
Helix        & 25   & $18.43$ & $7.14$ & $2.14$ \\
\hline
Lemniscate   & 25   & $19.81$ & $6.57$ & $2.37$ \\
\hline
Staircase    & 6.5  & $20.67$ & $5.14$ & $1.94$ \\
\hline
\end{tabular}
\end{center}
\end{table}

\subsection{Limitations}

The proposed framework is identified entirely from free-space motion. Consequently, the learned static and residual models do not explicitly account for external loading or contact interactions, and the resulting controller is expected to perform best within the sampled workspace, pressure limits, and motion bandwidth represented in the training data. Extending the approach to contact-rich manipulation will likely require training data collected under representative interaction conditions or explicit contact modeling.

Furthermore, the proposed stability analysis is local and relies on the learned linear residual dynamics. As such, it may not accurately predict behavior under large operating-point changes or long-term material drift.

Finally, this work considers a pneumatic bellows actuator as a representative soft robotic platform. While the proposed modeling framework is not inherently tied to pneumatic actuation, its applicability to other continuum robots, such as tendon-driven or concentric-tube systems, remains to be investigated. Evaluating the proposed control-oriented formulation across a broader range of soft robotic architectures is an important direction for future work.

\section{Conclusion}
\label{sec:conclusion}
This paper presents a control-oriented learning framework that achieves accurate dynamic control of soft pneumatic actuators without requiring high-fidelity physics-based models. By learning only the equilibrium behavior and transient residual dynamics, the proposed framework retains the simplicity of classical task-space feedback control while enabling local stability analysis directly from the learned model. Beyond accurate trajectory tracking, the same framework supports interactive user-guided motion and time-varying obstacle avoidance through reference modification, illustrating its flexibility for a range of robotic behaviors. More broadly, these results suggest that control-oriented learning provides a practical middle ground between physics-based and purely data-driven approaches, combining accurate closed-loop performance with computational efficiency, interpretability, and analytical tractability. We anticipate that this philosophy may provide a useful foundation for future data-driven control of increasingly complex soft robotic systems.

\section*{APPENDIX} \label{sec:appendix}

This appendix derives the augmented closed-loop system used for the local stability analysis in Section~\ref{sec:results_stab}. 

The normalized residual dynamics are given by (\ref{eq:edmdc}), where $\tilde{P}_k=\tilde{P}_{\mathrm{FF},k}+\tilde{P}_{\mathrm{FB},k}$. The feedback controller in (\ref{eq:fb}) can be written as:
\begin{equation}
P_{\mathrm{FB},k}
=
\tilde{K}_Pe_k
+
\tilde{K}_IE_k,
\end{equation}
where $\tilde{K}_P = M'K_P$, $\tilde{K}_I = M'K_I$, and $M'=M^T(MM^T+\alpha I)^{-1}$.
The normalized feedback input is:
\begin{equation}
\tilde{P}_{FB,k}
=
S_u^{-1}\tilde{K}_Pe_k
+
S_u^{-1}\tilde{K}_IE_k.
\label{eq:PFBnorm}
\end{equation}
Substituting (\ref{eq:PFBnorm}) into (\ref{eq:edmdc}) results in:
\begin{equation}
\tilde{z}_{k+1}
=
A\tilde{z}_k
+
B\tilde{P}_{FF,k}
+
BS_u^{-1}\tilde{K}_Pe_k
+
BS_u^{-1}\tilde{K}_IE_k.
\label{eq:dyn_appendix}
\end{equation}
The tracking error is $e_k = p_{\mathrm{ref},k}-p_k = p_{\mathrm{ref},k} - (p_{\mathrm{FF},k}+r_k) = p_\mu - CS_z\tilde{z}_k$, where $p_\mu = p_{\mathrm{ref},k} - p_{\mathrm{FF},k} - CS_z\mu_z$. Substituting this into (\ref{eq:dyn_appendix}):
\begin{align}
\tilde{z}_{k+1}
&=
A\tilde{z}_k
+
B\tilde{P}_{FF,k}
+
BS_u^{-1}\tilde{K}_P
(p_\mu-CS_z\tilde{z}_k)
+
BS_u^{-1}\tilde{K}_IE_k,
\\
&=
\left(
A
-
BS_u^{-1}\tilde{K}_PCS_z
\right)
\tilde{z}_k
+
BS_u^{-1}\tilde{K}_IE_k
+
p_\alpha,
\end{align}
where $p_\alpha = B\tilde{P}_{FF,k} + BS_u^{-1}\tilde{K}_Pp_\mu
$ contains exogenous feedforward and reference terms. Defining the augmented state:
\[
q_k=
\begin{bmatrix}
\tilde{z}_k\\
E_k
\end{bmatrix}
\in
\mathbb{R}^{(n_z+3)},
\]
and noting that $E_{k+1} = E_k+e_k = E_k+p_\mu-CS_z\tilde{z}_k$, the augmented closed-loop dynamics become:
\begin{align}
q_{k+1}
&=
\begin{bmatrix}
A-BS_u^{-1}\tilde{K}_PCS_z &
BS_u^{-1}\tilde{K}_I\\
-CS_z &
I
\end{bmatrix}
q_k
+
\begin{bmatrix}
p_\alpha\\
p_\mu
\end{bmatrix},
\\
&=
A_{\mathrm{aug}}q_k+g.
\label{eq:aug_appendix}
\end{align}
Since the affine term $g= \begin{bmatrix}
p_\alpha\\
p_\mu
\end{bmatrix}
$
does not affect the eigenvalues of the system, local closed-loop stability is governed solely by the augmented state transition matrix \(A_{\mathrm{aug}}\). Therefore, the learned closed-loop system is locally stable whenever:
\[
\rho(A_{\mathrm{aug}})<1,
\]
where \(\rho(\cdot)\) denotes the spectral radius.



\bibliographystyle{IEEEtran}
\bibliography{ref} 

\end{document}